\documentclass[10pt, a4paper]{article}
\usepackage{times}
\usepackage{booktabs}
\usepackage{multirow}
\usepackage{tikz}
\usepackage{tcolorbox}
\tcbuselibrary{breakable,skins}
\usepackage{graphicx}
\usetikzlibrary{shapes.geometric, arrows.meta, positioning}
\usepackage{lrec2026}
\usepackage{hyperref}
\usepackage{url}

\usepackage{fontspec}
\defaultfontfeatures{Ligatures=TeX}
\usepackage[bidi=default, english]{babel}
\babelprovide[import=fa]{persian}
\babelfont[persian]{rm}[Extension=.otf, UprightFont=*, BoldFont=*Bold, ItalicFont=*Italic]{FreeSerif}

\title{Culturally Adaptive Explainable LLM Assessment for Multilingual Information Disorder: A Human-in-the-Loop Approach}

\name{Maziar Kianimoghadam Jouneghani}
\address{University of Turin \\
        maziar.kianimoghadam@edu.unito.it}

\abstract{
Recognizing information disorder is difficult because judgments about manipulation depend on cultural and linguistic context. Yet current Large Language Models (LLMs) often behave as monocultural, English-centric “black boxes,” producing fluent rationales that overlook localized framing. Preliminary evidence from the multilingual Information Disorder (InDor) corpus suggests that existing models struggle to explain manipulated news consistently across communities. To address this gap, this ongoing study proposes a Hybrid Intelligence Loop, a human-in-the-loop (HITL) framework that grounds model assessment in human-written rationales from native-speaking annotators. The approach moves beyond static target-language few-shot prompting by pairing English task instructions with dynamically retrieved target-language exemplars drawn from filtered InDor annotations through In-Context Learning (ICL). In the initial pilot, the Exemplar Bank is seeded from these filtered annotations and used to compare static and adaptive prompting on Farsi and Italian news. The study evaluates span and severity prediction, the quality and cultural appropriateness of generated rationales, and model alignment across evaluator groups, providing a testbed for culturally grounded explainable AI. 
}

\begin{document}
\maketitleabstract

\section{Introduction and Motivation}
\label{sec:intro}
The proliferation of manipulated news, categorized as ``Information Disorder'' \citep{wardle2017information}, is deeply rooted in localized cultural and linguistic contexts. Historically, much of the NLP literature approached fake news detection as a binary classification problem, but there is an urgent need for Explainable AI (XAI) methods that provide natural language rationales \citep{shu2017fake,zhou2020survey,gurrapu2023rationalization,zhao2024explainability}. Furthermore, while Large Language Models (LLMs) possess powerful generative capabilities, they often operate as English-centric ``black boxes,'' exhibiting monocultural bias and cultural misalignment when evaluating non-English narratives \citep{tao-etal-2024-cultural,naous-etal-2024-beer}.

Recent baseline evaluations on the multilingual InDor corpus \citelanguageresource{InDor} provide preliminary evidence of this cross-cultural misalignment in current LLMs. Building directly on those foundational benchmarks, this ongoing study proposes a Human-in-the-Loop (HITL) \citep{monarch2021,wu-etal-2022-hitl} framework for culturally adaptive, explainable news assessment. More specifically, it introduces a culturally adaptive prompting setup that replaces static language-matched prompts with English task instructions and dynamically retrieved target-culture references via In-Context Learning (ICL).

In particular, to address the cross-cultural limitations identified in the InDor baselines, this study contributes:
\begin{itemize}
    \item A staged culturally adaptive HITL design that uses an Exemplar Bank seeded from filtered human-written InDor annotations, while reserving continuous expert correction for later phases.
    \item A prompting setup that replaces static language-matched few-shot examples with English task instructions and dynamically retrieved target-language exemplars.
    \item A pilot protocol for testing whether this design reduces cross-cultural gaps in LLM assessment through comparisons of span and severity prediction, rationale quality and cultural appropriateness, and alignment across evaluator groups in Farsi and Italian.
\end{itemize}

\section{Starting Point: Baseline Evidence of Cross-Cultural Misalignment}
\label{sec:baseline}
Foundational benchmarks on the InDor corpus \citelanguageresource{InDor}---which comprises 4,155 annotated news items across English, Farsi, Italian, and Russian---evaluating LLaMA 4 Maverick \citep{llama4} and Mixtral-8x22B \citep{mixtral} highlight limitations in current automated systems.

First, BERTScore \citep{zhang2019bertscore} evaluations showed that few-shot prompting improved span detection for target languages (e.g., Farsi, Russian) but degraded English performance for non-native evaluators \citelanguageresource{InDor}. The benchmark's few-shot templates use static, language-matched exemplars with English output tags (e.g., \texttt{<SPANS>}). While this approach generally improves performance, it can still fail when fixed exemplar narratives mismatch the test item's specific cultural frame, motivating our dynamic retrieval strategy. Furthermore, consistently low human Inter-Annotator Agreement (IAA) for span selection highlights the task's subjectivity and motivates a controlled exemplar-bank design based on filtered, article-level InDor annotations rather than conflicting annotation variants for the same item.

\begin{table}[h!]
    \centering
    \resizebox{\columnwidth}{!}{%
    \begin{tabular}{lcc}
    \toprule
        \textbf{Task} & \begin{tabular}[c]{@{}c@{}}\textbf{Gender}\\ \textbf{(M / F)}\end{tabular} & \begin{tabular}[c]{@{}c@{}}\textbf{Annotator L1}\\ \textbf{(Italian / Farsi)}\\ \textbf{[English items]}\end{tabular} \\
    \midrule
         Severity (F-Score) & 0.44 / 0.37 & 0.36 / 0.47 \\
         Rationale Agreement (1--4) & 3.28 / 2.68 & 2.44 / 3.53 \\
    \bottomrule
    \end{tabular}%
    }
    \caption{Alignment between LLaMA 4 outputs and human annotators, stratified by gender and annotator L1 evaluating English news (adapted from the InDor baseline \protect\citelanguageresource{InDor}).}
    \label{tab:english_eval}
\end{table}

Second, current LLMs exhibit a pronounced human-model alignment gap (Table \ref{tab:english_eval}). LLaMA 4 Maverick aligned more closely with male annotators and Farsi speakers than with female annotators and Italian speakers. These demographic disparities serve as a measurable symptom of a broader issue: standard models struggle to generalize their reasoning across diverse cultural lenses. Consequently, static prompt translation is likely insufficient to capture localized nuances and bridge these cultural divides, strongly motivating our dynamic, retrieval-based adaptive framework.

\section{Proposed Methodology: The Hybrid Intelligence Loop}
\label{sec:methodology}
To address the identified limitations, we propose an explainable adaptive assessment framework within a Human-in-the-Loop (HITL) setting. The full architecture combines exemplar-bank construction, culturally adaptive prompting, and a later expert-correction phase within a staged pipeline. In this study, we use the term \textit{rationale} to refer to the human-written explanation provided in an InDor annotation to justify why a selected span or news item is considered problematic; in the original InDor annotation setup, these explanations were written in free text and were often encouraged through an \textit{if \ldots then \ldots} structure \citelanguageresource{InDor}. However, the initial pilot implements only the seeding stage, allowing us to isolate the effect of dynamic exemplar retrieval while keeping the proof-of-concept operationally feasible. A detailed outline of this staged design is provided in Box~1.

\begin{tcolorbox}[colback=purple!4,
  colframe=purple!40!black, boxrule=0.8pt, rounded corners, title=\textbf{Box 1: Exemplar-Bank Construction and Revision}]
\small
\textbf{Phase 1: Corpus Mining (Seeding)}\\
Extract culturally representative InDor annotations that can serve as stable retrieval references while explicitly avoiding mismatched English-centric framings.

\vspace{1.5mm}
\hrule
\vspace{1.5mm}

\textbf{Phase 2: Active Error Correction}\\
Reviewing baseline LLM outputs to identify culturally misaligned rationales or misaligned severities, manually correcting the labels and rewriting the rationales.
\end{tcolorbox}

\subsection{Step 1: Problematic News Assessment}
The LLM processes multilingual news using the static baseline setups (B0: zero-shot target-language instructions; B1: static target-language few-shot) to generate an initial severity label, problematic text span, and rationale.

\subsection{Step 2: Expert Human Curation}
To build a dynamic \textbf{Multilingual Exemplar Bank} representing localized cultural knowledge, domain experts follow a two-phase curation protocol that specifies how culturally representative examples are selected, filtered, and refined before being used as in-context references.

Here, we use the term \textit{culturally misaligned rationale} to refer to a fluent explanation that is insufficiently grounded in the target cultural and discursive context and instead reflects monocultural or culturally blind assumptions \citep{tao-etal-2024-cultural,naous-etal-2024-beer,yari-koto-2025-unveiling}.

For the pilot reported here, exemplar-bank seeding is operationalized through a strict article-level filtering and deduplication pipeline applied to pre-existing InDor annotations.

\begin{enumerate}
    \item \textbf{Filtering Unusable Records:} Unusable records are removed, including N/A items and rows that cannot support prompting. For \textit{Problematic} annotations, this requires a valid severity label, a selected problematic span, and a written rationale.
    
    \item \textbf{Excluding Binary Conflicts:} We remove article-level binary conflicts, namely cases in which one annotation labels the same article as \textit{None} while another labels it as \textit{Problematic}. These cases are excluded at the article level and are not carried into the Exemplar Bank.
    
    \item \textbf{Resolving Severity Conflicts:} When multiple annotations mark the same article as \textit{Problematic} but differ in severity, span selection, or rationale wording, we retain only one article-level exemplar using a blind tie-breaking rule based on rationale character length. This prevents the bank from storing duplicate versions of the same news item with competing labels or alternative rationales. Because in-context learning is sensitive to the choice and composition of demonstrations \citep{zhang-etal-2022-active}, and because demonstration label words act as anchor-like references for final prediction \citep{wang2023label}, keeping a single representative exemplar helps reduce redundant or potentially conflicting guidance during retrieval-based prompting.
    
    \item \textbf{Test--Bank Split:} Finally, after this article-level deduplication step, the resulting clean resource is split into a held-out pilot test set and a retrieval bank so that no article can appear in both pools.
\end{enumerate}

We use rationale length as a simple tie-breaking criterion for three main reasons. First, it avoids manual selection and keeps the filtering process transparent. Second, in a subjective task like this one, longer rationales often provide more complete and informative explanations. Third, they tend to contain richer contextual detail, which may later support more effective retrieval of culturally relevant examples. Although length is not a direct measure of quality, it provides a practical and consistent heuristic when no adjudicated gold rationale is available.

In the full active-correction phase, domain experts follow a strict linguistic and cultural rubric. Rather than merely fact-checking claims, they evaluate the LLM sensitivity to implicit geopolitical framing, local idioms, and socio-historical tropes that characterize regional information disorder. For instance, an expert may revise a rationale that misses the mocking tone of a culturally specific metaphor or misidentifies a familiar local propaganda narrative.

To resolve disagreements, items are first reviewed independently and then discussed under a shared rubric. If disagreement persists, a third expert adjudicates the case; unresolved items are excluded from the Exemplar Bank.

In the full architecture, these corrected edge cases, together with mined examples, allow the repository to evolve toward the LLM's recurrent cross-cultural blind spots.

\subsection{Step 3: Culturally Adaptive Prompting}
To address cross-cultural mismatches, our adaptation mechanism leverages In-Context Learning (ICL) \citep{brown2020language,dong-etal-2024-icl,luo-etal-2024-retrieved-icl}. The framework replaces static target-language few-shot examples with $k$ dynamically retrieved same-language references from the Exemplar Bank.

When the LLM fails on a localized issue (e.g., misinterpreting Farsi sanctions news), the system queries the Exemplar Bank and injects contextually similar target-language references into the prompt. This retrieved context is intended to recalibrate the model's reasoning and anchor its assessment in a more appropriate cultural frame. More broadly, this design aligns with recent retrieval-augmented approaches showing that dynamically selected contextual evidence can improve alignment in culturally sensitive or misinformation-related settings without requiring task-specific fine-tuning \citep{seo-etal-2025-valuesrag,liu-etal-2025-raemollm}.

Structurally, the baseline setup (B1) uses target-language instructions paired with static target-language few-shot examples, whereas the adaptive framework replaces the static context with dynamically retrieved same-language references from the Exemplar Bank. Within this adaptive framework, M1 pairs the retrieved references with English task instructions, while A1 uses the same retrieved references with target-language instructions. Box~2 illustrates the contrast between the baseline and adaptive prompt structure using the Farsi pilot setup.

\newcommand{\placeholder}[1]{{\normalfont\small\textless #1\textgreater}}

% In the document
\begin{tcolorbox}[
  colback=purple!4,
  colframe=purple!40!black,
  boxrule=0.8pt,
  rounded corners,
  title=\textbf{Box 2: Prompt Architecture Comparison (Farsi Setting)},
  left=1mm,
  right=1mm,
  top=1mm,
  bottom=1mm,
  fontupper=\small\linespread{1.05}\selectfont
]
\textbf{\textcolor{red}{STATIC BASELINE (B1): Target-Language Instruction}}\par
\smallskip
\foreignlanguage{persian}{شما یک متخصص چارچوب‌بندی و سوگیری زبانی هستید. وظیفه شما تجزیه و تحلیل گزیده‌های خبری و شناسایی بازه‌های متنی است...}\par
\smallskip
\textit{[Translation:] You are a framing and language bias expert. Your job is to analyze news excerpts and identify text spans...}\par
\smallskip
\placeholder{Static Farsi Exemplars Injected Here}\par
\placeholder{Unseen Farsi News Item to Evaluate}

\medskip
\hrule
\medskip

\textbf{\textcolor{blue}{ADAPTIVE (M1): English Instruction + Target-Language Exemplars}}\par
\smallskip
You are a framing and language bias expert. Your job is to analyze news excerpts and identify text spans...\par
\smallskip
\placeholder{Dynamically Retrieved Farsi Exemplars Injected Here}\par
\placeholder{Unseen Farsi News Item to Evaluate}
\end{tcolorbox}

The task definition remains constant across conditions, but the instruction language is deliberately separated from the culturally grounded references.

To preserve relevance without conditioning on the unknown target label, retrieval relies on same-language semantic matching. For a target news item $x_{test}$ in language $\ell$, the system queries the language-specific subset $E_{\ell}$ of the Exemplar Bank $E$. We avoid exact-match filtering by severity, since the target severity is inherently unknown at inference time. Instead, dense semantic retrieval retrieves culturally relevant edge cases regardless of their labels. Strict deduplication ensures that no test article appears among the retrieved references.

Each candidate reference $e_i \in E_{\ell}$ and the query item $x_{test}$ are then encoded by a pretrained embedding function $f_{emb}$ (e.g., using sentence-transformer class models) and ranked by cosine similarity:

\begin{equation}
\mathrm{sim}(x_{test}, x_i) =
\frac{f_{emb}(x_{test}) \cdot f_{emb}(x_i)}
{\|f_{emb}(x_{test})\| \, \|f_{emb}(x_i)\|}
\end{equation}

The retrieval output is an ordered set of the top-$k$ references,
$R_k(x_{test}) = \{e_1, \dots, e_k\}$,
where each tuple
$e_i = (x_i, y_i^{span}, y_i^{sev}, r_i, m_i)$
contains the news text, span annotation, severity label, rationale, and metadata. For uncontested \textit{None} items, the span and rationale fields are explicitly passed as empty arrays (e.g., \texttt{[]}), signaling to the model that no manipulative framing is present. These retrieved references are then injected into the LLM context as in-context evidence, paired with English task instructions in M1 or target-language instructions in A1. Given the target item and the retrieved references, the model is prompted to produce the same structured output as in the baseline setting: a severity label, a problematic span, and a rationale.

At this stage, we do not impose fixed label quotas or artificial balancing constraints on the retrieved exemplars. This keeps the pilot focused on the effect of dynamic retrieval itself, without adding a second reranking mechanism that could blur the comparison with the static baseline. If later error analysis reveals systematic skew, such as majority-class dominance or recurrent false positives, a reranking strategy can be introduced as a separate ablation.

\begin{figure}[h!]
    \centering
    \resizebox{\columnwidth}{!}{
    \begin{tikzpicture}[
        node distance=2.2cm, auto,
        basebox/.style={rectangle, draw=black, fill=purple!5, text width=3.5cm, align=center, rounded corners, thick, minimum height=1.5cm},
        adaptbox/.style={rectangle, draw=black, fill=purple!10, text width=3.7cm, align=center, rounded corners, thick, minimum height=1.5cm},
        arrow/.style={draw=black, -{Latex[length=3mm, width=2mm]}, thick}
    ]
    
    % Titles
    \node [text width=3.5cm, align=center] (title_base) {\textbf{Baseline Prompt}};
    \node [text width=3.7cm, align=center, right of=title_base, node distance=4.3cm] (title_adapt) {\textbf{Culturally Adaptive Prompt}};
    
    % Row 1: Instructions
    \node [basebox, below of=title_base, node distance=1.2cm] (base_inst) {Target-Language\\Instructions};
    \node [adaptbox, below of=title_adapt, node distance=1.2cm] (adapt_inst) {English Task\\Instructions};
    
    % Row 2: Exemplars
    \node [basebox, below of=base_inst] (base_ex) {Static\\Target-Language\\Examples};
    \node [adaptbox, below of=adapt_inst] (adapt_ex) {Retrieved Target-Culture\\Exemplars};
    
    % Row 3: Target News
    \node [basebox, below of=base_ex] (base_news) {Target-Language\\News Item};
    \node [adaptbox, below of=adapt_ex] (adapt_news) {Target-Language\\News Item};
    
    % Arrows
    \path [arrow] (base_inst) -- (base_ex);
    \path [arrow] (base_ex) -- (base_news);
    
    \path [arrow] (adapt_inst) -- (adapt_ex);
    \path [arrow] (adapt_ex) -- (adapt_news);
    
    \end{tikzpicture}
    }
    \caption{Structural transition from the original baseline to the proposed Culturally Adaptive prompt. The adaptive approach resolves the reasoning and cultural framing mismatch through retrieved target-culture exemplars.}
    \label{fig:prompt_architecture}
\end{figure}
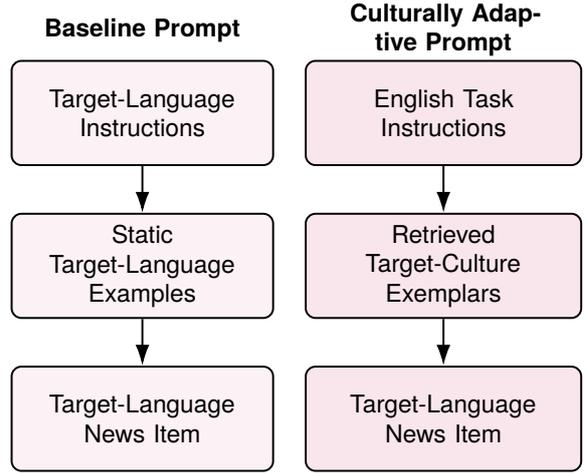

As shown in Figure \ref{fig:prompt_architecture}, \textit{baseline prompting} pairs target-language instructions with static language-matched few-shot examples. We hypothesize that instruction language impacts reasoning fidelity, and fixed exemplars fail to capture narrative-specific framing. Conversely, our adaptive approach pairs English task instructions---motivated by the hypothesis that they may better leverage the models' instruction-tuning for logical adherence \citep{shi2022language,ahuja2023mega}---with retrieved target-culture exemplars that ground the reasoning in locally relevant human references. Strict deduplication is used to prevent contamination between retrieved exemplars and test items.

\section{Pilot Evaluation Protocol}
\subsection{Scope and Experimental Setup}
\begin{figure}[h]
    \centering
    \resizebox{\columnwidth}{!}{
    \begin{tikzpicture}[
        node distance=1.9cm, auto,
        box/.style={rectangle, draw=purple!50!black, fill=purple!4, text width=7.5cm, align=center, rounded corners, minimum height=1.2cm, thick},
        arrow/.style={draw=purple!50!black, -{Latex[length=3mm, width=2mm]}, thick}
    ]
        % Nodes
        \node [box] (step1) {\textbf{1. Initial News Assessment}\\
        Standard LLM Generation\\
        (Potentially Culturally Misaligned)};
        \node [box, below of=step1] (step2) {\textbf{2. Expert Human Curation}\\Exemplar Bank Seeding \& Error Correction};
        \node [box, below of=step2] (step3) {\textbf{3. Culturally Adaptive Prompting}\\In-Context Learning (ICL) Injection};
        
        % Vertical Arrows
        \path [arrow] (step1) -- (step2);
        \path [arrow] (step2) -- (step3);
        
        % Feedback Loop Coordinates
        \coordinate (Loop_Left)  at ([xshift=-0.6cm]step1.west);
        \coordinate (Loop_Bottom) at ([yshift=-0.7cm]step3.south);
        
        % Feedback Loop Arrow
        \draw [arrow] (step3.south) 
            -- (step3.south |- Loop_Bottom) 
            -- (Loop_Left |- Loop_Bottom) 
            -- (Loop_Left |- step1.west) 
            -- (step1.west);
            
        % Text below the loop
        \node at ([yshift=-1.2cm]step3.south) {\textbf{\textit{Dynamically updates detection criteria}}};
        
    \end{tikzpicture}
    }
    \caption{The proposed Hybrid Intelligence Loop. The full architecture supports iterative expert feedback, while the initial pilot evaluates only the seeded Exemplar Bank phase before introducing continuous correction.}
    \label{fig:hitl_loop}
\end{figure}
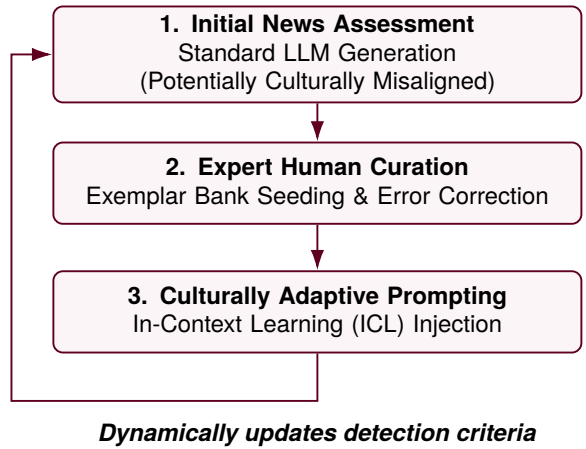

Building on prior InDor evaluations, we initiate a Phase~1 pilot to validate the proposed adaptive loop under a deliberately bounded evaluation design. As illustrated in Figure~\ref{fig:hitl_loop}, the full Hybrid Intelligence architecture supports a continuous expert feedback loop that dynamically updates detection criteria, whereas this initial pilot isolates the seeded Exemplar Bank phase in order to evaluate the core retrieval mechanism.

The pilot focuses on Farsi and Italian, the two language settings for which native-speaking evaluators are currently available, while Russian remains a planned extension if additional raters can be recruited. For each language, five native-speaking evaluators assess a mutually disjoint set of 20 news items, yielding exactly 100 uniquely evaluated items per language and 200 overall. This sample size was selected to balance the cognitive demands placed on human evaluators with the practical constraints imposed by the filtered corpus. Because the strict deduplication and conflict-resolution pipeline naturally reduces the pool of valid annotations in both Farsi and Italian, limiting the held-out pilot set to 100 items per language helps maintain a robust human evaluation while preserving a sufficiently diverse Exemplar Bank for effective semantic retrieval. Evaluator demographics (e.g., gender) are collected through consented self-report and used for exploratory subgroup analyses where sample size allows.

Because the pilot isolates the curation phase (Step~2 in Figure~\ref{fig:hitl_loop}), the Exemplar Bank is seeded directly from pre-existing InDor annotations rather than from a continuous expert correction loop. Before sampling the pilot items, the raw annotations are transformed into a clean article-level master table using the strict deduplication and filtering pipeline detailed in Section~3.2. This clean resource is then strictly divided into the retrieval bank and the held-out pilot test set to prevent data contamination.

This design preserves the natural class skew of the source corpus rather than forcing an artificial 50--50 balance in the bank. We retain uncontested \textit{None} items in the cleaned article pool, but we do not require a fixed \textit{None}/\textit{Problematic} ratio in each prompt. Instead, the bank remains naturalistic, while experimental control is maintained through the held-out evaluation design and the fixed retrieval procedure.

To ensure reproducibility, we access all generations via the OpenRouter API\footnote{\url{https://openrouter.ai/}} and retain the baseline model families evaluated in the foundational InDor benchmarks \citelanguageresource{InDor}. Specifically, we use LLaMA~4 Maverick \citep{llama4} (a 400B-parameter Mixture-of-Experts model with a 1.05M-token context) and Mixtral-8x22B-Instruct \citep{mixtral} (a 141B-parameter MoE with a 65.5K-token context). While the original InDor study used a decoding temperature of 1.0, we do not compare our results directly to those aggregate scores. Instead, we re-run all prompting conditions on our held-out pilot set using temperature 0.0. This choice reduces generation variability and helps keep the comparison focused on differences in prompt design and retrieval setup rather than sampling randomness.

\begin{figure}[h]
    \centering
    \resizebox{\columnwidth}{!}{
    \begin{tikzpicture}[
        node distance=0.6cm and 0.4cm,
        cond/.style={
            circle,
            draw=purple!50!black,
            fill=purple!4,
            thick,
            minimum size=1.45cm,
            font=\Large\bfseries
        },
        inst/.style={
            rectangle,
            draw=purple!50!black,
            fill=white,
            rounded corners,
            thick,
            text width=4.0cm,
            align=center,
            minimum height=1.5cm,
            font=\large
        },
        eninst/.style={
            rectangle,
            draw=blue!60!black,
            fill=blue!5,
            rounded corners,
            thick,
            text width=4.0cm,
            align=center,
            minimum height=1.5cm,
            font=\large
        },
        ex/.style={
            rectangle,
            draw=purple!50!black,
            fill=purple!10,
            rounded corners,
            thick,
            text width=5.1cm,
            align=center,
            minimum height=1.5cm,
            font=\large
        },
        noex/.style={
            rectangle,
            draw=gray!70,
            fill=gray!5,
            rounded corners,
            thick,
            text width=5.1cm,
            align=center,
            minimum height=1.5cm,
            font=\large\itshape,
            text=gray!70
        },
        plus/.style={
            font=\Huge\bfseries,
            text=purple!50!black
        }
    ]

    % Row 1: B0
    \node[cond] (b0) {B0};
    \node[inst, right=of b0] (b0_inst) {Target-Language\\Instructions};
    \node[plus, right=of b0_inst, node distance=0.15cm] (p1) {+};
    \node[noex, right=of p1, node distance=0.15cm] (b0_ex) {Zero-Shot};

    % Row 2: B1
    \node[cond, below=of b0] (b1) {B1};
    \node[inst, right=of b1] (b1_inst) {Target-Language\\Instructions};
    \node[plus, right=of b1_inst, node distance=0.15cm] (p2) {+};
    \node[ex, right=of p2, node distance=0.15cm] (b1_ex) {Static\\Language-Matched Few-Shot};

    % Row 3: M1
    \node[cond, below=of b1] (m1) {M1};
    \node[eninst, right=of m1] (m1_inst) {\textbf{English}\\Task Instructions};
    \node[plus, right=of m1_inst, node distance=0.15cm] (p3) {+};
    \node[ex, right=of p3, node distance=0.15cm] (m1_ex) {Retrieved Target-Language\\Cultural Exemplars};

    % Row 4: A1
    \node[cond, below=of m1] (a1) {A1};
    \node[inst, right=of a1] (a1_inst) {Target-Language\\Instructions};
    \node[plus, right=of a1_inst, node distance=0.15cm] (p4) {+};
    \node[ex, right=of p4, node distance=0.15cm] (a1_ex) {Retrieved Target-Language\\Cultural Exemplars};

    \end{tikzpicture}
    }
    \caption{Visual breakdown of the four prompting conditions used in the pilot. M1 combines English task instructions with retrieved target-language cultural exemplars, while A1 keeps the same retrieval setup but switches the instruction language back to the target-language.}
    \label{fig:prompt_conditions}
\end{figure}
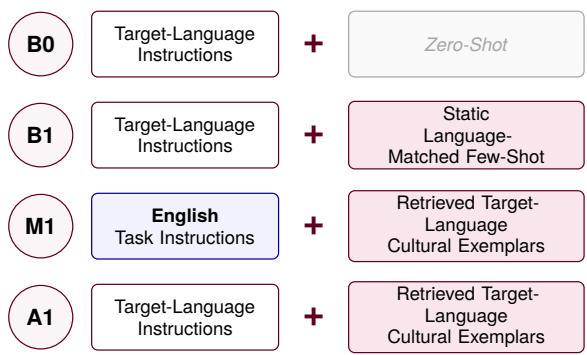

As illustrated in Figure~\ref{fig:prompt_conditions}, we test four distinct prompting conditions designed to cleanly isolate the effects of contextual exemplars, adaptive retrieval, and instruction language. The zero-shot baseline (\textbf{B0}) provides the model with only target-language instructions. The few-shot static baseline (\textbf{B1}) represents the foundational InDor setup, pairing target-language instructions with static, language-matched few-shot examples. Our proposed adaptive method (\textbf{M1}) replaces this static context by pairing English task instructions with dynamically retrieved, same-language exemplars from the seeded Exemplar Bank. Finally, to isolate the effect of the retrieval mechanism from the instruction language, we include an instruction ablation (\textbf{A1}) that combines target-language instructions with the exact same dynamically retrieved exemplars used in M1.

To reduce rating bias, the downstream evaluation uses a blinded A/B design administered through a structured questionnaire platform such as Qualtrics. Evaluators see the original news item together with paired, anonymized, and randomized rationales generated by the static baseline and the adaptive method. They rate each rationale on overall quality as well as three sub-dimensions: factual grounding in the text, sensitivity to cultural nuance, and avoidance of generic English-centric assumptions.

Before evaluation, native-speaking raters receive a short rubric defining a culturally misaligned rationale as an explanation that is insufficiently grounded in the target cultural-political context, misses locally salient framing, or relies on an inappropriate generic interpretive frame.

\subsection{Operational Metrics}
We evaluate the framework using a combination of automated and human-centered metrics:

\begin{itemize}
    \item \textbf{Automated Evaluation (Severity, Span, and Rationale):} Before human evaluation, LLM generations are assessed automatically. Because our data preparation pipeline (Section 3.2) resolves annotation conflicts to yield a single reference per article, we evaluate directly against these filtered labels to enable a stable comparison between static and adaptive prompting:
    \begin{itemize}
        \item \textit{Severity Classification} is measured via Macro-F1. While the original InDor baseline evaluated models against distinct demographic subgroups to examine subjective alignment patterns, our pilot evaluates predictions against a deduplicated article-level reference in order to support a consistent comparison across prompting conditions.
        \item \textit{Span Detection} is evaluated using best-match token-overlap F1. The foundational InDor study used exact token-match accuracy to estimate human agreement and BERTScore for generative span evaluation. In our setting, we adopt token-overlap F1 because exact matching can be overly strict for generated spans, while semantic similarity metrics do not directly evaluate boundary accuracy. Token-overlap F1 therefore provides a more suitable balance between strictness and partial credit. For uncontested \textit{None} items, where the ground-truth span is empty, the F1 score is strictly defined as 1.0 if the model correctly outputs an empty array, and 0.0 if it incorrectly extracts a text span.
        \item \textit{Rationale Quality} is assessed via BERTScore \citep{zhang2019bertscore} to measure the semantic similarity between the generated explanation and the corresponding human-written rationale. This follows the general logic of the original InDor evaluation, while remaining appropriate for explanations that may differ lexically but convey similar meaning.
    \end{itemize}

    \item \textbf{Rationale Cultural Appropriateness (Human Evaluation):} Native-speaking evaluators participate in a blinded A/B test, rating paired rationales (static vs.\ adaptive) on a 1--4 Likert scale using a brief rubric. A rationale is treated as culturally misaligned when it is insufficiently grounded in the target cultural and discursive context and instead relies on monocultural or culturally blind assumptions \citep{tao-etal-2024-cultural,naous-etal-2024-beer,yari-koto-2025-unveiling}. The rubric asks whether the rationale (i) is grounded in the text, (ii) captures the locally relevant manipulative framing, and (iii) avoids generic or misplaced interpretive assumptions.

    \item \textbf{Alignment Disparity:} To examine whether dynamic retrieval reduces the alignment gaps observed in the original InDor baselines, we calculate performance differences across demographic subgroups (e.g., gender or language) as $\Delta = |F1_{Group A} - F1_{Group B}|$ on severity prediction. Comparing $\Delta_{baseline}$ and $\Delta_{adapted}$ indicates whether the framework reduces variation in model alignment across evaluator groups.
\end{itemize}

\section{Expected Findings and Limitations}
\label{sec:expected}
\subsection{Hypothesized Outcomes}
We anticipate that the adaptive framework will yield an increase in Rationale Cultural Appropriateness. Qualitative case studies are expected to show a transition from generic justifications to more contextualized explanations reflecting local sociopolitical realities. Furthermore, we hypothesize a decreased Alignment Disparity, suggesting a reduction in demographic sensitivities. In the pilot, the goal is not to demonstrate full-scale deployment, but to test whether a seeded exemplar bank already provides measurable gains under a cost-conscious setup. An example of this expected qualitative shift is presented in Box 3.

\begin{tcolorbox}[
  enhanced,
  breakable,
  colback=purple!4,
  colframe=purple!55!black,
  boxrule=0.8pt,
  rounded corners,
  title=\textbf{Box 3: Hypothesized Shift in Rationale Generation in Farsi},
  left=1mm,
  right=1mm,
  top=1mm,
  bottom=1mm,
  fontupper=\small\linespread{1.05}\selectfont
]
\textbf{Context:} A Farsi news article attributing domestic economic challenges entirely to foreign sanctions using politically charged framing.\par

\medskip

\textbf{\textcolor{red}{STATIC BASELINE (B1) Rationale (Baseline Output):}}\par
\smallskip
\foreignlanguage{persian}{"اگر مقاله داده‌های اقتصادی تأیید شده‌ای برای حمایت از ادعاهای خود ارائه ندهد، آنگاه مشکل‌ساز و سوداگرانه است."}\par
\smallskip
\textit{[Translation:] ``If the article does not provide verified economic data to support its claims, then it is problematic and speculative.''}\par
\smallskip
\textbf{Critique:} Relies on a generic Western journalistic heuristic (source verification) while missing the actual mechanics of state-aligned economic propaganda.

\medskip
\hrule
\medskip

\textbf{\textcolor{blue}{ADAPTIVE (M1) Rationale (Culturally Grounded):}}\par
\smallskip
\foreignlanguage{persian}{"اگر مقاله مسائل پیچیده اقتصادی داخلی را صرفاً به بازیگران خارجی نسبت دهد، آنگاه با تکیه بر یک سپر بلای ژئوپلیتیک آشنا، خواننده را از عوامل سیاست داخلی منحرف می‌کند."}\par
\smallskip
\textit{[Translation:] ``If the article attributes complex domestic economic issues solely to external foreign actors, then it distracts the reader from internal policy factors by relying on a familiar geopolitical scapegoat.''}\par
\smallskip
\textbf{Critique:} Better captures the locally salient manipulative framing anchored by the retrieved target-culture exemplars.
\end{tcolorbox}

Specifically, we expect the adaptive framework to reduce generic, template-like LLM judgments that fail to capture local media realities. For example, when analyzing Farsi news, baseline models often rely on standard Western journalistic heuristics, such as checking for ``verified sources.'' However, in state-controlled media environments, a cited official source may itself be a vehicle for propaganda; thus, treating source verification as a proxy for neutrality completely misses the underlying ideological bias \citelanguageresource{InDor}. Conversely, in the Italian context, problematic discourse is rarely about state censorship, but is instead frequently characterized by specific sociopolitical framing, such as manipulating narratives around immigration, European geopolitical relations, or institutional conflicts (e.g., framing the judiciary in opposition to the government) \citelanguageresource{InDor}. We hypothesize that the M1 adaptive setup will shift the LLM's focus away from generic fact-checking and toward recognizing these localized tactics. By anchoring the reasoning in expert-curated examples, the system should better detect specific manipulative framing devices---whether state-aligned economic scapegoating in Farsi or biased immigration narratives in Italian---that native speakers intuitively recognize.

\subsection{Risks and Limitations}
The primary limitation of this framework is the potential for model overfitting to the subjective viewpoints represented in the Exemplar Bank. To mitigate the risk of optimizing toward specific annotator-group norms rather than a single stable reference, the pilot relies on high-quality pre-existing InDor annotations and a balanced downstream evaluation design. For the full architecture, preventing long-term subjectivity drift during the continuous expert correction cycle will require explicit governance mechanisms, including periodic inter-annotator agreement audits, rotating expert review, and agreement-based admission rules so that no single annotator can unilaterally introduce new exemplars into the bank. By testing exemplar-bank seeding first, this pilot deliberately isolates the effect of retrieval-based ICL on culturally grounded reasoning, leaving the evaluation of a continuous, resource-intensive expert correction cycle for future work.

A second limitation concerns language coverage. The initial pilot focuses only on Farsi and Italian, since these are the language settings in which balanced native-speaker evaluation is currently feasible. Russian remains part of the broader architecture but is deferred until sufficient evaluators become available. More broadly, the generalizability of this approach to additional languages and language families remains an open research question.

We also deliberately avoid several alternative design choices in the pilot. First, we do not preserve multiple conflicting versions of the same article inside the Exemplar Bank, because the bank is meant to provide stable adaptive references rather than model the full spread of human disagreement. Second, we do not rebalance the bank to an artificial 50--50 label distribution, since this would move the retrieval resource away from the natural distribution of the source corpus. Third, we do not force each prompt to contain a preset mixture of \textit{None} and \textit{Problematic} exemplars. Starting from pure same-language semantic top-$k$ retrieval keeps the main comparison methodologically clean; if a retrieval-skew problem is later observed, it can be addressed in a separate ablation with empirical justification.

Furthermore, while the current phase of the Hybrid Intelligence Loop focuses on retrieval-based adaptation through In-Context Learning (ICL), the architecture also suggests a possible longer-term extension beyond inference-time prompting. As the Multilingual Exemplar Bank grows through iterative expert curation, it may eventually serve not only as a retrieval resource but also as a high-quality source of culturally grounded supervision. One possible future research direction would be to investigate whether periodically audited expert corrections could support lightweight instruction tuning or post-training alignment. In such a setup, fine-tuning could help the model absorb more stable cross-cultural reasoning patterns into its parameters, while dynamic ICL would still provide narrative-specific grounding at inference time.

\section{Conclusion}
\label{sec:conclusion}
The multifaceted nature of information disorder requires detection systems that are as culturally nuanced as the narratives they analyze. This ongoing study argues that monocultural, English-centric LLMs remain limited in their ability to provide culturally grounded and explainable assessments in multilingual settings. In response, it proposes an adaptive human-in-the-loop framework that combines expert-curated exemplar retrieval with LLM generation. Rather than relying on static prompting alone, the methodology investigates whether culturally adaptive prompting can improve span and severity assessment while producing more culturally appropriate natural-language rationales. In this way, the study positions human cultural knowledge not as a post hoc correction, but as a central component of multilingual explainable AI.

\section*{Ethical Considerations}
\label{sec:ethics}
This research involves the analysis of the Information Disorder (InDor) corpus, which contains real-world instances of disinformation, misinformation, and malinformation. Consequently, the dataset includes content that may be considered offensive, encompassing racist, sexist, or violent language, as well as polarizing ideological rhetoric.

Comprehensive ethical guidelines regarding the collection, annotation, and handling of this sensitive content are detailed in the primary InDor dataset study \citelanguageresource{InDor}. Our downstream evaluation will adhere to these established ethical protocols. To mitigate risks within our HITL framework, the Exemplar Bank strictly anonymizes annotator IDs and requires cross-checking by multiple domain experts before adding entries. This expert curation helps ensure that the stored human rationales remain grounded in the item and consistent with annotation guidelines, preserving context while reducing the risk of amplifying highly polarized or toxic views.

\section*{Acknowledgements}
I would like to express my deepest gratitude to my supervisor, Prof. Viviana Patti, and to Marco Antonio Stranisci for their invaluable guidance, unwavering support, and kindness.

Furthermore, I dedicate this study, its theoretical design, and all future implementations of this system---regardless of whether they ultimately succeed or fail in their empirical results---to the people of Iran. The full framework, and the effort behind it, is for you.

\renewcommand{\refname}{\vspace{-2ex}}
\section*{Bibliographical References}
\label{sec:reference}
\bibliographystyle{lrec2026-natbib}
\bibliography{references}
\section*{Language Resource References}
\label{lr:ref}
\bibliographystylelanguageresource{lrec2026-natbib}
\bibliographylanguageresource{languageresource}

\end{document}